\documentclass{article}

\usepackage{arxiv}

\usepackage[utf8]{inputenc} 
\usepackage[T1]{fontenc}    
\usepackage{hyperref}       
\usepackage{url}            
\usepackage{booktabs}       
\usepackage{amsfonts}       
\usepackage{nicefrac}       
\usepackage{microtype}      
\usepackage{lipsum}		
\usepackage{graphicx}

\usepackage{doi}

\usepackage[utf8]{inputenc} 
\usepackage[T1]{fontenc}    
\usepackage{hyperref}       
\usepackage{url}            
\usepackage{booktabs}       
\usepackage{amsfonts}       
\usepackage{nicefrac}       

\usepackage{xcolor}         
\usepackage{amsbsy}
\usepackage[ruled]{algorithm2e} 
\usepackage{amsmath,epsfig}
\usepackage{amsfonts,amssymb,amsbsy,textcomp,marvosym,amsmath,caption,threeparttable,amsthm,subfigure,float,lastpage,lscape}
\usepackage{eurosym,mathrsfs,fancyhdr,CJK,multicol,graphics,indentfirst,color,bm,upgreek,booktabs,graphicx,multirow,warpcol}
\usepackage{epstopdf}
\usepackage{amssymb}
\usepackage{amsmath}
\usepackage{graphicx}
\usepackage{multirow}

\usepackage{amssymb}
\title{A Transductive Maximum Margin Classifier for Few-Shot Learning}

\author{ Fei Pan\\
	National Key Lab for Novel Software Technology\\
	Nanjing University\\
	Nanjing, 210023 \\
	\texttt{felix.panf@outlook.com} \\
	\And
	Chunlei Xu\\
	National Key Lab for Novel Software Technology\\
	Nanjing University\\
	Nanjing, 210023 \\
	\texttt{xu.chunlei@outlook.com} \\
	\And
	Jie Guo\\
	National Key Lab for Novel Software Technology\\
	Nanjing University\\
	Nanjing, 210023 \\
	\texttt{guojie@nju.edu.cn} \\
	\And
	Yanwen Guo\\
	National Key Lab for Novel Software Technology\\
	Nanjing University\\
	Nanjing, 210023 \\
	\texttt{ywguo@nju.edu.cn} \\
}

\hypersetup{
pdftitle={A Transductive Maximum Margin Classifier for Few-Shot Learning},
pdfsubject={},
pdfauthor={Fei Pan et al.},
pdfkeywords={Few-shot learning, maximum margin classifier, separating hyperplane, L-BFGS},
}

\begin{document}
\maketitle

\begin{abstract}
	Few-shot learning aims to train a classifier that can generalize well when just a small number of labeled examples per class are given. We introduce a transductive maximum margin classifier for few-shot learning (FS-TMMC). The basic idea of the classical maximum margin classifier is to solve an optimal prediction function so that the training data can be correctly classified by the resulting classifer with the largest geometric margin. In few-shot learning, it is challenging to find such classifiers with good generalization ability due to the insufficiency of training data in the support set. FS-TMMC leverages the unlabeled query examples to adjust the separating hyperplane of the maximum margin classifier such that the prediction function is optimal on both the support and query sets. Furthermore, we use an efficient and effective quasi-Newton algorithm, the L-BFGS method for optimization. Experimental results on three standard few-shot learning benchmarks including miniImagenet, tieredImagenet and CUB show that our method achieves state-of-the-art performance.
\end{abstract}


\renewcommand{\thefootnote}{}
\footnotetext{Preprint. Work in progress.}
\section{Introduction}

In recent years, deep learning models have made remarkable progress in various vision and language tasks \cite{ResNet,ren2015faster,devlin2018bert}. However, they have a common limitation, that training effective deep neural networks requires a great number of labeled examples. Besides, it is costly to collect and label the large amount of training data. In some situations, it is extremely difficult to obtain enough training data, such as medical image processing, scarce objects identification. When the training data is insufficient, the trained model is prone to overfitting. Notably, humans are capable of learning from a small number of examples by leveraging past experiences. Inspired by this, few-shot learning, which is designed to learn a new visual concept through a few training examples, is attracting more and more research attentions.

In order to alleviate the overfitting problem caused by the inadequate labeled data in few-shot learning, Vinyals {\it et al.} \cite{MatchingNet} propose an episodic training strategy to learn from diverse tasks over vast episodes. In each episode, the algorithm learns from a few labeled examples in the support set, which will be used to make predictions in the unlabeled query examples. This learning strategy simulates the evaluation phase where there are insufficient labeled examples as well as some unlabeled data. The consistency between the training and evaluation phases reduces generalization errors by alleviating the distribution gap \cite{TPN}. It has been adopted by many approaches in few-shot learning \cite{MAML, REPTILE, LEO,LGMNet,kim2019edge}. However, this strategy cannot resolve the problem of learning from inadquate training data.

\newcommand{\mysize}{0.40\textwidth}
\begin{figure*}[!t]
\centering
  \subfigure[]{
    \includegraphics[width=\mysize]{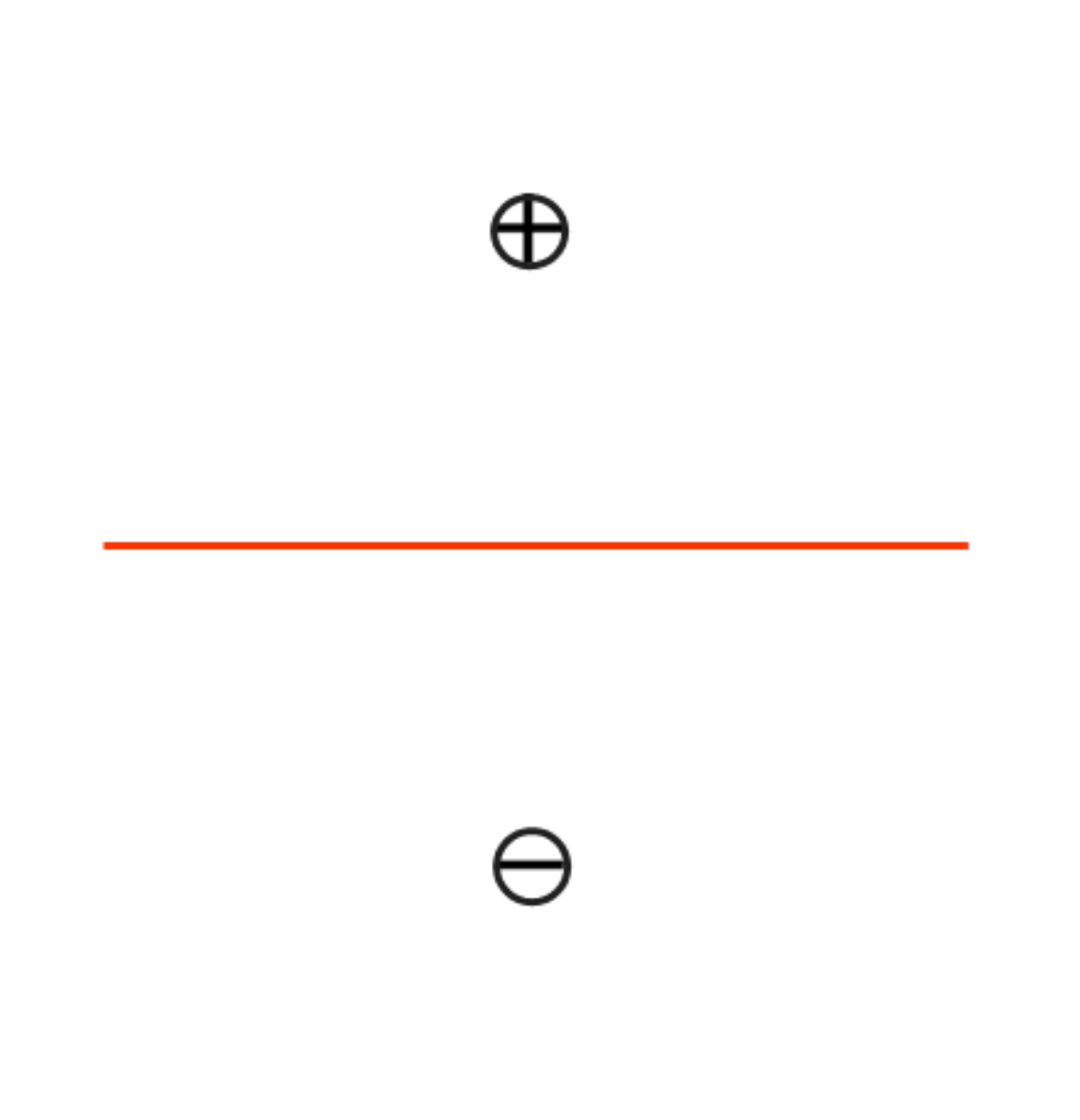}}
  \quad
  \subfigure[]{
    \includegraphics[width=\mysize]{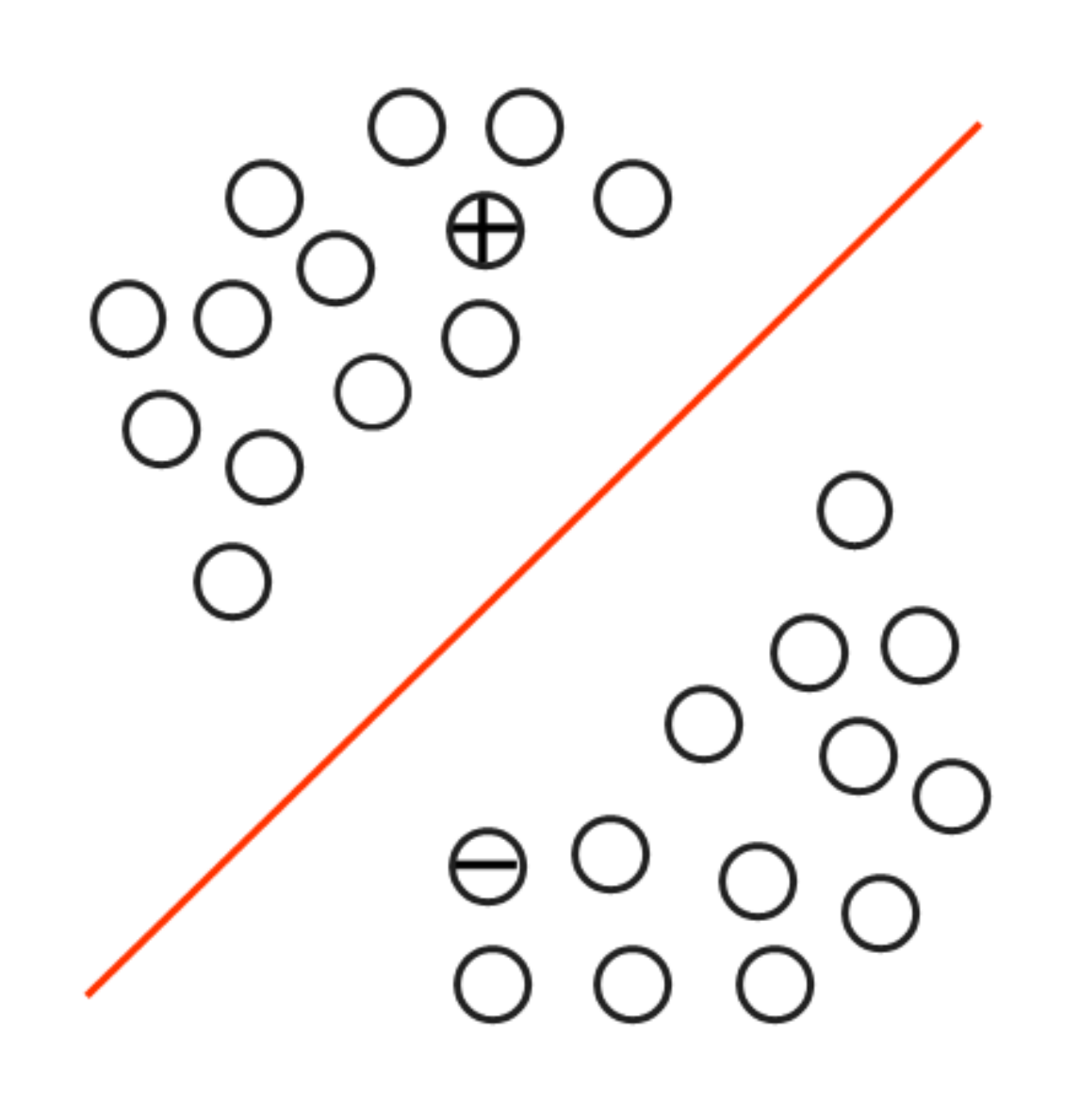}}
  \caption{An example of the 2-way 1-shot problem where there are totally two categories and each category contains one example in the support set. The symbols \textcircled{+}, \textcircled{-}, \textcircled{ } in the figure denote the labeled positive example, the labeled negative example and the unlabeled example, respectively. (a) The maximum margin classifier is solved with the support examples. (b) The query examples are also involved in solving the maximum margin classifier. }\label{fig1}
\end{figure*}

In this paper, we propose a transductive maximum margin classifier for few-shot learning (FS-TMMC). The classical maximum margin classifier aims to search for a predicton function so that the corresponding separating hyperplane correctly divides the training data and the resulting classifier has the largest geometric margin. In few-shot learning scenarios, the training examples are scarce, not enough to find a separating hyperplane that generalizes well on unseen data. FS-TMMC is constructed using a mixture of the labeled support examples and the unlabeled query examples in a given task. Different from the maximum margin classifier, the objective of FS-TMMC also involves assigning category labels to the query examples when constructing the maximum margin classifier. In order to obtain the labels of the query examples, we calculate the prediction function on the query set by taking advantage of both the support and query examples. 

The involvement of the query examples in finding the optimal prediction function brings improvement to the maximum margin classifier because of the insufficiency of the training data in the support set. The unlabeled examples in the query set would adjust the separating hyperplane such that the prediction function is optimal on both the support and query examples. Please refer to Figure~\ref{fig1} for an example that illustrates the research motivation of FS-TMMC. Given the support examples only, the maximum margin classifier would produce an optimal separating hyperplane as illustrated by the solid red line in Figure~\ref{fig1}(a). As shown in Figure~\ref{fig1}(b), the unlabeled examples in the query set are also involved in solving the optimal prediction function of the maximum margin classifier. We observe that the normal vector of the resulting separating hyperplane in Figure~\ref{fig1}(b) is distinct from that in Figure~\ref{fig1}(a). Besides, the solved optimal prediction function in Figure~\ref{fig1}(a) would produce a few misclassifications in the query set.

Our paper makes three main contributions. First, we propose a transductive maximum margin classifier for few-shot learning (FS-TMMC) which adjusts the separating hyperplane of the maximum margin classifier by taking account of information from the query examples. Second, we leverage the one-vs-rest strategy to make our problem compatible with the binary classification model and use Platt scaling to transform the output of the corresponding prediction function into a probability distribution over categories. Third, we evaluate our method on three benchmark datasets for few-shot learning, namely miniImageNet, tieredImageNet and CUB. The experimental results show that FS-TMMC achieves state-of-the-art results.

\section{Related work}
There are a variety of works on Few-Shot Learning which can be broadly categorized into five categories by prior knowledge representations.

\noindent\textbf{Metric-based methods.} Koch {\it et al.} \cite{Siamese} propose the Siamese Network to measure the similarity between the query and the support examples. Snell {\it et al.} \cite{ProtoNet} propose to use the class prototypes which are calculated by averaging the embeddings of the support examples from each category and make predictions based on the distances of the query example with respect to all prototypes. Triantafillou {\it et al.} \cite{tieredImagenet} leverage the unlabeled examples when producing prototypes.

\noindent\textbf{Memory-based methods.} Santoro {\it et al.} \cite{MANN} introduce MANN which originally uses memory networks to store feature and label pairs, and update elements with the least recently used strategy. MetaNet \cite{MetaNet} stores model parameters in memory which are the fast weights for the support examples to help query examples get relevant information, and updates the parameters across tasks. 

\noindent\textbf{Optimization-based methods.} MAML \cite{MAML} learns good initialization of the parameters in the model. The model with the learned parameters may not perform well on the other tasks, but using these parameters as a starting point, it would learn from new tasks fast. 
Instead of adapting the high-dimensional model parameters through back propagation as in MAML \cite{MAML}, LEO \cite{LEO} learns the parameters of an encoder and an decoder, and makes adaptations in the low-dimensional latent space.

\noindent\textbf{Generation-based methods.} RTA \cite{RTA} produces parameters of neural networks based on the support examples in a task. But it only produces a part of the model parameters to adjust the network for the given task. LGM-Net \cite{LGMNet} directly generates the parameters of the matching networks based on the latent representations of the examples in the support set. 

\noindent\textbf{Transductive methods.} These methods construct the models using both the labeled and unlabeled examples, making full use of task information. TPN \cite{TPN} learns to propagate labels by label propagation, and the weight matrix is obtained from learning. CAN-T \cite{CANT} utilizes attention mechanisms to calculate the cross attention maps between the support and the query examples, making the extracted features more discriminant.

Similar to our method, several recent approaches deal with the problem of few-shot learning without episodic training. They train general classification networks on the base dataset, which will be used as feature extractors in the evaluation stage. SimpleShot \cite{SimpleShot} shows that a simple nearest-neighbor classifier with feature transformations and L2-normalization leads to competitive few-shot learning results. LaplacianShot \cite{LaplacianShot} introduces a transductive inference algorithm that minimizes a Laplacian-regularized binary-assignment function to encourage similar examples to be assigned with the same labels. In contrast to the two approaches, our method leverages the examples in the query set to adjust the separating hyperplane of the maximum margin classifier.

\section{Main Approach}
\subsection{Problem Setting}
In few-shot image classification, there are three datasets with non-overlapping categories: the base set $\mathcal{D}_{base}$, 
the validation set $\mathcal{D}_{validation}$ and the test set $\mathcal{D}_{test}$. We train a general image classification model using the base set $\mathcal{D}_{base}$, which extracts the feature representations of the examples in the test set $\mathcal{D}_{test}$. The validation set $\mathcal{D}_{validation}$ is used to tune the hyper-parameters. In the evaluation stage of $N$-way $K$-shot tasks, we randomly sample $N$ categories from the test set $\mathcal{D}_{test}$ and $K$ examples from each category to construct the support set. Our approach is evaluated in the query set where $Q$ examples are provided for each category.

\subsection{FS-TMMC}
We begin by introducing the hinge loss function: $l(u)=\max(0, 1-u)$.
Support Vector Machines is the corresponding maximum margin classifier, which solves:
\begin{equation}\label{eq01}
\mathop{{\rm minimize}}_{f\in \mathcal{H}}\ \frac{\lambda}{2}\Vert f\Vert_\mathcal{H}^2+\frac{1}{n}\sum_{i=1}^{n}l(y_i f(\boldsymbol{x}_i)),
\end{equation}
where $f$ lies in a Reproducing Kernel Hilbert Space $\mathcal{H}$, $y_i$ belongs to $\{-1,+1\}$, and $\lambda \ (\lambda>0)$ is a hyper-parameter. 

We first train a base classification model with deep neural networks $g_{\boldsymbol{\theta}}$ parameterized by $\boldsymbol{\theta}$ using the base set $\mathcal{D}_{base}$. The output dimension of $g_{\boldsymbol{\theta}}$ is set to $D$. Then we sample $N$-way $K$-shot tasks from the test set $\mathcal{D}_{test}$. Each task contains a support set $\mathcal{S}=\{(Z_i,y_i)\}_{i=1}^{NK}$ and a query set $\mathcal{Q}=\{(Z_i,y_i)\}_{i=1}^{NQ}$. We leverage $g_{\boldsymbol{\theta}}$ to extract the compact feature representations of $\mathcal{S}$ and $\mathcal{Q}$ to obtain the pre-processed support set $\mathcal{S}^{\prime}=\{(\boldsymbol{z}_i,y_i) |  \boldsymbol{z}_{i}=g_{\boldsymbol{\theta}}(Z_i)$, $(Z_i,y_i)\in \mathcal{S}, i=1,\cdots,NK\}$ and the pre-processed query set  $\mathcal{Q}^{\prime}=\{(\boldsymbol{z}_i,y_i) |  \boldsymbol{z}_{i}=g_{\boldsymbol{\theta}}(Z_i)$, $(Z_i,y_i)\in \mathcal{Q}, i=1,\cdots,NQ\}$. Besides, we define the following matrices:
\begin{align}
	\boldsymbol{X}_{\mathcal{S}^{\prime}}&=[\boldsymbol{x}_1, \cdots, \boldsymbol{x}_{NK}],\\ \boldsymbol{x}_i&=\boldsymbol{z}_i,\ (\boldsymbol{z}_i,y_i)\in \mathcal{S}^{\prime},\ i\in\{1,\cdots,NK\};\nonumber\\
	\boldsymbol{X}_{\mathcal{Q}^{\prime}}&=[\boldsymbol{x}_{NK+1}, \cdots, \boldsymbol{x}_{NK+NQ}],\\ \boldsymbol{x}_{NK+i}&=\boldsymbol{z}_i,\ (\boldsymbol{z}_i,y_i)\in \mathcal{Q}^{\prime},\ i\in\{1,\cdots,NQ\}.\nonumber
\end{align}

We use the one-vs-rest strategy to make our problem compatible with the binary classification model. For each category $c\in\{1,\cdots,N\}$, we introduce a label vector of the support set $\boldsymbol{y}^{c}=[y_1^c, \cdots, y_{NK}^c]^T\in R^{NK}$, where the value of $y_i^c$ is set to $+1$ if $y_i$ equals $c$ and $-1$ otherwise, $(\boldsymbol{z}_i,y_i)\in\mathcal{S}^{\prime},i\in\{1,\cdots,NK\}$. 

Due to the introduction of the one-vs-rest strategy, the training examples of each category become unbalanced. So we assign category weights to all examples in the support set which are inversely proportional to category frequencies. The weight vector is $\boldsymbol{w}^c=[w_1^c,w_2^c,\cdots,w_{NK}^c]^T\in R^{NK}$. 

In addition to the labeled data $\boldsymbol{X}_{\mathcal{S}^{\prime}}$,  we also take advantage of the unlabeled examples $\boldsymbol{X}_{\mathcal{Q}^{\prime}}$ to get more accurate prediction functions \cite{bennett1999semi, gieseke2012sparse}. Specifically, we are intended to find the prediction function $f^c\in \mathcal{H}$ as well as the label assignments for the query examples $\boldsymbol{y}^{c\prime}=[y^{c\prime}_1, \cdots, y^{c\prime}_{NQ}]^T$ by the following optimization problem:

\begin{align}\label{eq02}
	\mathop{{\rm minimize}}_{f^c\in \mathcal{H},\boldsymbol{y}^{c\prime}} \ &\frac{\lambda_1}{2}\Vert f^c\Vert_\mathcal{H}^2\nonumber+\frac{1}{NK}\sum_{i=1}^{NK}w_i^c l(y_i^c  f^c(\boldsymbol{x}_i))\nonumber\\
	&+\lambda_2\frac{1}{NQ} \sum_{i=1}^{NQ}l(y^{c\prime}_i f^c(\boldsymbol{x}_{NK+i})),
\end{align}
where $\lambda_1,\lambda_2\ (\lambda_1,\lambda_2>0)$ are hyper-parameters.

The optimal value of $y^{c\prime}_i$ equals $1$ if $f^c(\boldsymbol{x}_{NK+i})\geq 0$ and $-1$ otherwise \cite{chapelle2005semi}. Thus we get the following optimization problem:
\begin{align}\label{eq04}
	\mathop{{\rm minimize}}_{f^c\in \mathcal{H}} \ &\frac{\lambda_1}{2}\Vert f^c\Vert_\mathcal{H}^2+\frac{1}{NK}\sum_{i=1}^{NK}w_i^c \max(0,1-y_i^c  f^c(\boldsymbol{x}_i))\nonumber\\&+\lambda_2\frac{1}{NQ} \sum_{i=1}^{NQ}\max(0,1-|f^c(\boldsymbol{x}_{NK+i})|).
\end{align}

Obviously, Eq.~(\ref{eq04}) contains the maximum and absolute value operations which are not differentiable. In order to optimize it efficiently, we replace the non-differentiable parts in Eq.~(\ref{eq04}) with their differentiable surrogates. Specifically, we use the LogSumExp function to approximate the maximum operator in the term $\max(0,1-y_i^c  f(\boldsymbol{x}_i))$: $\max(a_1,\cdots,a_n)\approx\frac{1}{\gamma_1}\log\sum_{i=1}^{n}\exp{(\gamma_1 a_n)}$ \cite{log-sum-exp}. Following Chapelle {\it et al.} \cite{chapelle2005semi}, we replace the term $\max(0,1-|f^c(\boldsymbol{x}_{NK+i})|)$ by the term ${\rm exp}(-\gamma_2 f^c(\boldsymbol{x}_{NK+i})^2)$. $\gamma_1,\gamma_2$ are used as hyperparameters.

According to the representer theorem \cite{sch2001a}, the solution of Eq.~(\ref{eq04}) satisfies 
\begin{equation}\label{eq03}
	f^{c*}(\boldsymbol{x})=\sum_{i=1}^{M}\alpha^{c*}_i k(\boldsymbol{x}_i,\boldsymbol{x}),
\end{equation}
where the function $k$ is the reproducing kernel of $\mathcal{H}$ and $M$ equals $NK\!+\!NQ$. By introducing the vector $\boldsymbol{\alpha}^c=[\alpha_1^c, \cdots, \alpha_{M}^c]^T$, $\boldsymbol{\alpha}^{c*}$ solves

\begin{align}\label{eq05}
	&\arg\min_{\boldsymbol{\alpha}^c} F(\boldsymbol{\alpha}^c) = \frac{\lambda_1}{2}(\boldsymbol{\alpha}^c)^T \boldsymbol{K}\boldsymbol{\alpha}^c\nonumber\\&+\frac{1}{NK}\sum_{i=1}^{NK}\frac{w_i^c}{\gamma_1}\log(1+{\rm exp}(\gamma_1(1-y_i^c  \sum_{j=1}^{M}\alpha^{c}_j k(\boldsymbol{x}_j,\boldsymbol{x}_i)))) \nonumber\\
	&+\lambda_2\frac{1}{NQ} \sum_{i=1}^{NQ}{\rm exp}(-\gamma_2 (\sum_{j=1}^{M}\alpha^{c}_j k(\boldsymbol{x}_j,\boldsymbol{x}_{NK+i}))^2),
\end{align}
where $\boldsymbol{K}$ is a matrix with elements $K_{ij}=k(\boldsymbol{x}_i, \boldsymbol{x}_j)$, $i,j\in\{1,2,\cdots, M\}$. 

The gradient of $F(\boldsymbol{\alpha}^c)$ w.r.t. $\boldsymbol{\alpha}^c$ is
\begin{equation}\label{eq06}
	\nabla_{\boldsymbol{\alpha}^c}F(\boldsymbol{\alpha}^c) = \boldsymbol{K}(\lambda_1\boldsymbol{\alpha}^c+\boldsymbol{t}),
\end{equation}
where $\boldsymbol{t}\in{R}^{M}$ is a vector with elements
\begin{equation}\label{eq07}
    t_i = \left\{
    \begin{aligned}
        &-\frac{o_i}{p_i}     & i\leq NK, \\
        &-\frac{q_i}{NQ}      & otherwise,
    \end{aligned}
    \right.
\end{equation}
and
\begin{align}
o_i&=w_i^c y_i^c{\rm exp}(\gamma_1(1-y_i^c \sum_{j=1}^{M}\alpha^{c}_j k(\boldsymbol{x}_j,\boldsymbol{x}_{i})))\nonumber,\\
p_i&=NK(1+{\rm exp}(\gamma_1(1-y_i^c \sum_{j=1}^{M}\alpha^{c}_j k(\boldsymbol{x}_j,\boldsymbol{x}_{i}))))\nonumber,\\
q_i&=2\gamma_2 \lambda_2 \sum_{j=1}^{M}\alpha^{c}_j k(\boldsymbol{x}_j,\boldsymbol{x}_{i}){\rm exp}(-\gamma_2 (\sum_{j=1}^{M}\alpha^{c}_j k(\boldsymbol{x}_j,\boldsymbol{x}_{i}))^2).\nonumber
\end{align}

\subsection{Optimization}
In order to optimize the proposed model efficiently, we use the BFGS algorithm which generates an approximate Hessian matrix at each step with a small computational cost. The iterative sequence generated by the approximate matrix rather than the Hessian matrix still has the property of superlinear convergence.

Given an initial $\boldsymbol{\alpha}_0^c\in{R}^{M}$, the BFGS algorithm calculates the sequence $\boldsymbol{\alpha}_1^c,\cdots,\boldsymbol{\alpha}_k^c,\boldsymbol{\alpha}_{k+1}^c$ iteratively until convergence via
\begin{equation}
\boldsymbol{\alpha}_{k+1}^c=\boldsymbol{\alpha}_k^c+t_k\boldsymbol{d}_k
\end{equation}
where $t_k$ is the stepsize and the direciton $\boldsymbol{d}_k=-\boldsymbol{H}^k \nabla_{\boldsymbol{\alpha}^c}F(\boldsymbol{\alpha}_k^c)$. 

The update of the inverse Hessian approximation is given by \cite{nocedal2006numerical}:\label{eq08}
\begin{equation}
    \boldsymbol{H}^{k+1}=(I-\rho_k \boldsymbol{s}_k \boldsymbol{w}_k^T)\boldsymbol{H}^k (I-\rho_k \boldsymbol{w}_k \boldsymbol{s}_k^T)+\rho_k\boldsymbol{s}_k\boldsymbol{s}_k^T
\end{equation}
where $\rho_k=(\boldsymbol{w}_k^T\boldsymbol{s}_k)^{-1}$, $\boldsymbol{w}_k=\nabla_{\boldsymbol{\alpha}^c}F(\boldsymbol{\alpha}_{k+1}^c)-\nabla_{\boldsymbol{\alpha}^c}F(\boldsymbol{\alpha}_k^c)$
and $\boldsymbol{s}_k=\boldsymbol{\alpha}_{k+1}^c-\boldsymbol{\alpha}_k^c$.

Although the BFGS method overcomes the difficulty of calculating the Hessian matrix, it still cannot be applied to large-scale optimization problems. Generally speaking, the inverse Hessian approximation $\boldsymbol{H}^k$ is a dense matrix, and storing the dense matrix consumes $O(n^2)$ memory, which is obviously impossible for large-scale problems. In this paper, we use the limited memory version of BFGS method (L-BFGS) to reduce the memeory storage as well as the computational cost \cite{gieseke2012sparse}.

When the value of $\lambda_2$ equals $0$ in Eq.~(\ref{eq05}), it corresponds to a special form of maximum margin classifer. In order to adjust the separating hyperplane of the maximum margin classifier gradually, we initially set the value of $\lambda_2$ to $0$ and increase its value step by step. This way of updating the value of $\lambda_2$ results in a sequence $\boldsymbol{v}_{\lambda_2}=\{v_0,v_1,v_2,\cdots\}$.

When the iteration of $\boldsymbol{v}_{\lambda_2}$ ends and the termination criteria of L-BFGS method is satisfied, we get the output values of $f^{c*}$ evaluated on $\boldsymbol{x}_i$ ($i\in\{1,\cdots,M\}$) with Eq.~(\ref{eq03}) : $e_i^c=f^{c*}(\boldsymbol{x}_i)$. Then, we use Platt scaling \cite{john1999prob} to transform the predicted values $e_i^c$ ($i\in\{1,\cdots,NK\}$) into a probability distribution over categories. Specifically, we firstly set $s_i^c$ to $0$ if $y_i^c$ equals $-1$ and $+1$ otherwise; then we leverage the data pairs $\{(e_i^c, s_i^c)\}_{i=1}^{NK}$ to train a logistic regression model $LR^c$. The predicted probabilities of the examples on the query set are given by $p_j^c=LR^c(e_{NK+j}^c)$, $j=1,\cdots,NQ$. Finally, we obtain the predicted labels of the query examples: $\hat{y}_j=\arg\max_c p_j^c$, $j=1,\cdots,NQ$. The overall method is summerized in Algorithm~\ref{alg1}.

\begin{algorithm}[H]\label{alg1}
  \SetAlgoLined
  \KwIn{the base dataset $\mathcal{D}_{base}$, the test dataset $\mathcal{D}_{test}$,
		feature extractor $g_{\boldsymbol{\theta}}$ with parameter $\boldsymbol{\theta}$,
		the hyper-parameters: $\lambda_1, \gamma_1, \gamma_2$ and $\boldsymbol{v}_{\lambda_2}=\{v_0,v_1,v_2,\cdots \}$
	}

  Optimize $g_{\boldsymbol{\theta}}$ using the base set $\mathcal{D}_{base}$\;
  Sample a batch of $N$-way $K$-shot tasks $\{\mathcal{T}_i\}_{i=1}^T$ from the test set $\mathcal{D}_{test}$\;
  \For{each $\mathcal{T}_i$}
	{
		Obtain the pre-processed sets $\mathcal{S}^{\prime}=\{(\boldsymbol{z}_i,y_i)\}_{i=1}^{NK}$ and $\mathcal{Q}^{\prime}=\{(\boldsymbol{z}_i,y_i)\}_{i=1}^{NQ}$\;
		\For{$c$ in $\{1,\cdots,N\}$}
		{
			Obtain the vectors $\boldsymbol{y}^c$, $\boldsymbol{w}^c$ and $\boldsymbol{x}_i, i=1,\cdots,M$\;
			\For {$\lambda_2$ in $\{v_0,v_1,v_2,\cdots \}$}
			{
				Initialize the inverse Hessian approximation $\boldsymbol{H}^0$\;
				Initialize $\boldsymbol{\alpha}_0^c$  if $\lambda_2$ equals 0\;
				$k=0$\;
				\While{not converged}{
					Obtain $\nabla_{\boldsymbol{\alpha}^c}F(\boldsymbol{\alpha}_k^c)$ using Eq.~(\ref{eq06})\;
					Calculate $\boldsymbol{d}_k=-\boldsymbol{H}^k \nabla_{\boldsymbol{\alpha}^c}F(\boldsymbol{\alpha}_k^c)$\;
					Find the appropriate step size $t_k$ by line search and update $\boldsymbol{\alpha}_{k+1}^c=\boldsymbol{\alpha}_k^c+t_k\boldsymbol{d}_k$\;
					Update the inverse Hessian approximation $\boldsymbol{H}^{k+1}$ using L-BFGS\;
					$k=k+1$\;
				}
				$\boldsymbol{\alpha}_0^c=\boldsymbol{\alpha}_k^c$\;
			}
			
			Acquire $p_j^c$ $(j\in\{1,\cdots,NQ\})$ with Platt Scaling\;
		}
		Obtain the predicted labels on the query set $\hat{y}_j=\arg\max_c p_j^c$, $j=1,\cdots,NQ$.
	}
  \caption{FS-TMMC}
\end{algorithm}

\section{Experiment} \label{experiment}
\subsection{Datasets}
We experiment on three benchmarks for few-shot classification: miniImagenet, tieredImagenet and CUB-200-2011.

\noindent\textbf{miniImagenet.} The miniImagenet dataset \cite{MatchingNet} is constructed from the ImageNet \cite{ImageNet} dataset, which is the largest image recognition database in the world to facilitate the study of visual recognition. The miniImagenet dataset consists of 100 categories and 600 images of size 84$\times$84 per category. Following the split in LaplacianShot \cite{LaplacianShot}, we divide the dataset into the base, validation and test sets, with 64, 16, and 20 categories respectively.

\noindent\textbf{tieredImagenet.} The tieredImagenet \cite{tieredImagenet} is a larger subset of the ImageNet dataset, containing totally 608 categories. Some categories are grouped together to form a parent category. There are 34 parent categories, each with about 10 to 30 categories. Following the splits introduced by Triantafillou {\it et al.} \cite{tieredImagenet}, we use 351 categories for the base set, 97 categories for the validation set and 160 categories for the test set. 

\noindent\textbf{CUB-200-2011}. The CUB-200-2011 \cite{CUB} is a bird image classification dataset, which is also the benchmark in the research of fine-grained classification and recognition. Following LaplacianShot \cite{LaplacianShot}, we split it into three sets where there are 100 categories for the base set, 50 categories for the validation set and 50 categories for the test set. We resize the images to $84\times84$ pixels. 

\subsection{Network Models}
We adopt three different network architectures as the feature extractors, namely ResNet, WRN and DenseNet. 

\noindent\textbf{ResNet} \cite{ResNet} introduces shortcut connections to neural networks, making deeper neural networks easier to optimize. We take advantage of a variant of the standard 18-layer ResNet architecture with 8 basic residual blocks, where we remove the first two down-sampling layers. Besides, we set the stride and the kernel size in the first convolutional layer to 1 and $3$, respectively.  

\noindent\textbf{WRN} \cite{WRN} uses a shallower but wider model to effectively improve the performance of ResNet. Following the architecture used in LEO \cite{LEO}, we set the number of convolutional layers and the widening factor in WRN to 28 and 10, respectively. 

\noindent\textbf{DenseNet} \cite{DenseNet} directly connects all layers under the premise of ensuring the maximum information transmission between layers in neural networks.
We adopt the standard 121-layer architecture with 58 dense blocks, leaving out the first two down-sampling operations and setting the kernel size in the first convolutional layer to $3$. 

\subsection{Evaluation Protocol}
Following \cite{LEO, SimpleShot}, we evaluate on 5-way 1-shot and 5-way 5-shot tasks by randomly sampling 10,000 tasks from the test set. Each $N$-way $K$-shot task has $K$ images per category in the support set and $Q$ $(Q=15)$ images per category in the query set. We report the averaged accuracy of 10,000 tasks with the 95\% confidence interval in the following experiments.

\subsection{Implementation Details}\label{Details}
\noindent\textbf{Base model training.} The feature extractor is trained by the standard cross-entropy loss function on the base set, with a label-smoothing of parameter 0.1. The model is optimized by SGD, with the learning rate initialized to 0.1. We multiply the learning rate by 0.1 every 30 epoches. During training, we use the data augmentation strategies including random cropping, color jitter
and random horizontal flipping. We resize all the images to 84 $\times$ 84 during training. Besides, we set the batch size to 256, 128 and 100, for ResNet, WRN and DenseNet, respectively. The models are trained with four GeForce GTX 1080Ti GPUs.

\noindent\textbf{Feature transformation.} 
During evaluation of a task sampled from test set, we first compute the mean of the features, $\bar{\boldsymbol{z}}=\frac{1}{|\mathcal{S}^{\prime}|+|\mathcal{Q}^{\prime}|}\sum_{(\boldsymbol{z}_i,y_i)\in\mathcal{S}^{\prime} \bigcup \mathcal{Q}^{\prime} } \boldsymbol{z}_i$.
Then, we center the image features by subtracting this mean: $\boldsymbol{z}=\boldsymbol{z}-\bar{\boldsymbol{z}}$. At last, the image features are normalized by L2 normalization: $\boldsymbol{z}=\frac{\boldsymbol{z}}{\|\boldsymbol{z}\|_2}$.

\begin{table*}[htbp]
	\centering
	\caption{Averaged accuracy (in \%) in miniImageNet and tieredImageNet. The values represent the averaged accuracies in 10,000 episodes from the test set with 95\% confidence intervals. The best results are reported in bold font.}
	\begin{tabular}{cccccc}
		\toprule 
		\multirow{2}{*}{Methods} & \multirow{2}{*}{Network} & \multicolumn{2}{c}{miniImageNet} & \multicolumn{2}{c}{tieredImageNet}    \\
		\cline{3-6}
		& &  5-way 1-shot & 5-way 5-shot & 5-way 1-shot & 5-way 5-shot \\
		\midrule
		{MAML}\cite{MAML} & ResNet-18 & 49.61$\pm$0.92 &65.72$\pm$0.77& - & -\\
		{Chen et al.}\cite{chen} & ResNet-18 & 51.87$\pm$0.77 &75.68$\pm$0.63 & - & -\\
		
		{RelationNet}\cite{RelationNet}&ResNet-18&52.48$\pm$0.86&69.83$\pm$0.68 & - & -\\
		
		{MatchingNet}\cite{MatchingNet}&ResNet-18&52.91$\pm$0.88&68.88$\pm$0.69 & - & -\\
		
		{Gidaris et al.}\cite{Gidaris} & ResNet-15 &55.45$\pm$0.89& 70.13$\pm$0.68 & - & -\\
		{ProtoNet}\cite{ProtoNet} & ResNet-18 &54.16$\pm$0.82&73.68$\pm$0.65 & - & -\\
		{SNAIL}\cite{SNAIL} & ResNet-15 &55.71$\pm$0.99&68.88$\pm$0.92 & - & -\\
		{Bauer et al.}\cite{Bauer} & ResNet-34 &56.30$\pm$0.40&73.90$\pm$0.30 & - & -\\
		{AdaCNN}\cite{adaCNN} & ResNet-15 &56.88$\pm$0.62&71.94$\pm$0.57 & - & -\\
		{TADAM}\cite{TADAM} & ResNet-15 &58.50$\pm$0.30&76.70$\pm$0.30 & - & -\\
		{CAML}\cite{CAML} & ResNet-12 &59.23$\pm$0.99&72.35$\pm$0.71 & - & -\\
		{TPN}\cite{TPN} & ResNet-12 &59.46&75.64 & - & -\\
		{TEAM}\cite{TEAM} & ResNet-18 &60.07&75.90 & - & -\\
		{MTL}\cite{MTL} & ResNet-18 &61.20$\pm$1.80&75.50$\pm$0.80 & - & -\\
		{VariationalFSL}\cite{VariationalFSL} & ResNet-18 &61.23$\pm$0.26&77.69$\pm$0.17 & - & -\\	
		{Transductive\ tuning}\cite{Transductivetuning} & ResNet-12 &62.35$\pm$0.66&74.53$\pm$0.54 & - & -\\	
		{MetaoptNet}\cite{MetaoptNet} & ResNet-18 &62.64$\pm$0.61&78.63$\pm$0.46 & 65.99$\pm$0.72 & 81.56$\pm$0.53\\
		{SimpleShot}\cite{SimpleShot} & ResNet-18 &63.10$\pm$0.20&79.92$\pm$0.14 & 69.68$\pm$0.22 & 84.56$\pm$0.16\\
		{CAN+T}\cite{CANT} & ResNet-12 &67.19$\pm$0.55&80.64$\pm$0.35 & 73.21$\pm$0.58 &84.93$\pm$0.38\\
		{LaplacianShot}\cite{LaplacianShot} & ResNet-18 &\textbf{72.11}$\pm$0.19 & 82.31$\pm$0.14 & 78.98$\pm$0.21 & 86.39$\pm$0.16\\
		{FS-TMMC (ours)}  & ResNet-18  & 71.96$\pm$0.26 & \textbf{82.94}$\pm$0.15 & \textbf{79.67}$\pm$0.26 & \textbf{87.60}$\pm$0.15\\
		\cline{1-6}
		{Qiao}\cite{Qiao} & WRN &59.60$\pm$0.41&73.74$\pm$0.19 & - & -\\
		{LEO}\cite{LEO} & WRN &61.76$\pm$0.08&77.59$\pm$0.12 & 66.33$\pm$0.05 &81.44$\pm$0.09\\
		{ProtoNet}\cite{ProtoNet} & WRN &62.60$\pm$0.20&79.97$\pm$0.14 & - & -\\
		{CC+rot}\cite{CCrot} & WRN &62.93$\pm$0.45&79.87$\pm$0.33 & 70.53$\pm$0.51 & 84.98$\pm$0.36\\
		{MatchingNet}\cite{MatchingNet} & WRN &64.03$\pm$0.20&76.32$\pm$0.16 & - & -\\
		{FEAT}\cite{FEAT} & WRN &65.10$\pm$0.20&81.11$\pm$0.14 & 70.41$\pm$0.23& 84.38$\pm$0.16\\
		{Transductive\ tuning}\cite{Transductivetuning} & WRN &65.73$\pm$0.68&78.40$\pm$0.52 & 73.34$\pm$0.71& 85.50$\pm$0.50\\
		{SimpleShot}\cite{SimpleShot} & WRN &65.87$\pm$0.20&82.09$\pm$0.14 & 70.90$\pm$0.22 &85.76$\pm$0.15\\
		{SIB}\cite{SIB} & WRN &70.0$\pm$0.6&79.2$\pm$0.4 & - & -\\
		{BD-CSPN}\cite{BDCSPN} & WRN &70.31$\pm$0.93&81.89$\pm$0.60 & 78.74$\pm$0.95& 86.92$\pm$0.63\\							
		{LaplacianShot}\cite{LaplacianShot} & WRN &74.86$\pm$0.19 & 84.13$\pm$0.14 & 80.18$\pm$0.21 &87.56$\pm$0.15\\
		{FS-TMMC (ours)}  & WRN  &  \textbf{75.28}$\pm$0.24 & \textbf{85.06}$\pm$0.14 &\textbf{81.23}$\pm$0.25&\textbf{88.72}$\pm$0.15\\
		\cline{1-6}
		{SimpleShot}\cite{SimpleShot} & DenseNet &65.77$\pm$0.19&82.23$\pm$0.13 & 71.20$\pm$0.22 &86.33$\pm$0.15\\
		{LaplacianShot}\cite{LaplacianShot} & DenseNet &75.57$\pm$0.19 & 84.72$\pm$0.13 & 80.30$\pm$0.22 &87.93$\pm$0.15\\
		{FS-TMMC (ours)}  & DenseNet  &  \textbf{76.06}$\pm$0.25 & \textbf{85.73}$\pm$0.13 & \textbf{82.65}$\pm$0.25 & \textbf{89.84}$\pm$0.14 \\
		\bottomrule
	\end{tabular}
	\label{table1}
\end{table*}

\begin{table*}[htbp]
	\centering
	\caption{Averaged accuracy (in \%) in CUB and the cross-domain experimental setting from miniImagenet to CUB.}
	\begin{tabular}{cccccc}
		\toprule 
		\multirow{2}{*}{Methods} & \multirow{2}{*}{Network} & \multicolumn{2}{c}{CUB} & \multicolumn{2}{c}{miniImageNet $\longrightarrow$ CUB}    \\
		\cline{3-6}
		& &  5-way 1-shot & 5-way 5-shot & 5-way 1-shot & 5-way 5-shot \\
		\midrule	
		{MatchingNet}\cite{MatchingNet}&ResNet-18&73.49&84.45 & - & 53.07\\
		
		{MAML}\cite{MAML} & ResNet-18 &68.42 &83.47 & - & 51.34\\
		{ProtoNet}\cite{ProtoNet} & ResNet-18 &72.99&86.64&-&62.02\\
		{RelationNet}\cite{RelationNet}&ResNet-18&68.58&84.05&-&57.71\\
		{Chen et al.}\cite{chen} & ResNet-18 & 67.02 &83.58&-&65.57\\
		
		{SimpleShot}\cite{SimpleShot} & ResNet-18 &70.28&86.37&48.56&65.63\\
		{LaplacianShot}\cite{LaplacianShot} & ResNet-18 &80.96&88.68&\textbf{55.46}&66.33\\
		\cline{1-6}
		{FS-TMMC (ours)}  & ResNet-18  &  \textbf{81.53} & \textbf{89.61} & 54.75 & \textbf{69.72}\\
		\bottomrule
	\end{tabular}
	\label{table2}
\end{table*}

\begin{table*}[htb]
	\centering
	\caption{Ablation study of the involvement of the unlabeled examples in the query set in solving the optimal prediction function of the maximum margin classifier.}
	\begin{tabular}{cccccc}
		\toprule 
		\multirow{2}{*}{Network} & \multirow{2}{*}{Methods} & \multicolumn{2}{c}{miniImageNet} & \multicolumn{2}{c}{tieredImageNet}    \\
		\cline{3-6}
		& &  5-way 1-shot & 5-way 5-shot & 5-way 1-shot & 5-way 5-shot \\
		\midrule
		ResNet-18 & FS-MMC & 64.15  &79.79 & 71.83  & 85.18\\
		ResNet-18 & FS-TMMC & 71.96 & 82.94 & 79.67 & 87.60\\
		\midrule
		WRN & FS-MMC & 67.21  & 82.36 & 73.16  & 86.37 \\
		WRN & FS-TMMC &  75.28 & 85.06 &81.23 &88.72\\
		\midrule
		DenseNet & FS-MMC & 67.11 & 82.68 & 73.81 & 87.36 \\
		DenseNet & FS-TMMC &  76.06 & 85.73 & 82.65 & 89.84\\
		\midrule
	\end{tabular}
	\label{table3}
\end{table*}

\noindent\textbf{Experimental settings.} For simplicity, we keep the hyper-parameters of our method fixed across all the experiments which are obtained on the validation set of the miniImagenet dataset. The hyper-parameters are set as follows: $\lambda_1=0.04$, $\gamma_1=20.0$, $\gamma_2=2.0$. The sequence $\boldsymbol{v}_{\lambda_2}$ equals $\{0, 0.00001, 0.001, 0.1, 1.0\}$. Besides, we use the linear kernel function in all the experiments: $k(\boldsymbol{x}_i, \boldsymbol{x}_j)=\boldsymbol{x}_i^T \boldsymbol{x}_j$, $i,j\in\{1,2,\cdots, M\}$.

\subsection{Results}
We first compare our method with state-of-the-art approaches on miniImagenet and tieredImagenet datasets in the 5-way 1-shot and 5-way 5-shot experimental settings of few-shot learning. As shown in Table~\ref{table1}, FS-TMMC outperforms state-of-the-art
methods except for the 5-way 1-shot setting with the base model of ResNet-18 on miniImagenet where the accuracy is slightly lower than that of LaplacianShot \cite{LaplacianShot}. It's worth noting that our method outperforms all other methods in the tieredImagenet dataset across different network architectures and the largest performance gains over LaplacianShot are obtained when we use DenseNet as the base model.

Besides, we report the experimental results on CUB dataset in Table~\ref{table2}. Following Chen {\it et al.} \cite{chen}, we also conduct a cross-domain experiment from miniImagenet to CUB. Specifically, we
train a ResNet-18 model on the base set of miniImagenet and employ the trained model to extract features in the test set of CUB. As we can see, FS-TMMC outperforms the other methods by a large margins in the 5-way 5-shot cross-domain setting.

The inference run-time per few-shot learning task in the 5-way 5-shot setting on mini-ImageNet dataset with the base model of WRN is about 1.0 seconds on the Intel Core i9-9900X CPU.

\subsection{Ablation study}
In order to evaluate the effectiveness of the proposed method, we set the value of $\lambda_2$ to $0$ in Eq.~(\ref{eq05}) which is equivalent to the setting that the sequence $\boldsymbol{v}_{\lambda_2}=\{0\}$. In this case, the additional unlabeled examples in the query set are absent when solving the maximum margin classifer. The hyper-parameters are set as follows: $\lambda_1=0.04, \gamma_1=20.0$. In this experiment, the hyper-parameter $\gamma_2$ has no effect on the results, so we omit its value here. We call the resulting model FS-MMC in the following description. 

Table~\ref{table3} summerizes the results in this experimental setting compared with FS-TMMC. We observe that FS-TMMC outperforms FS-MMC by more than $7\%$ in 5-way 1-shot setting and $2\%$ to $3\%$ in 5-way 5-shot setting. The gains are consistent across different network architectures. As a consequence, we conclude that the involvement of the unlabeled query examples in solving the optimal prediction function of the maximum margin classifier plays a significant role in achieving better few-shot learning results.

\section{Conclusions}
In this paper, we propose a transductive maximum margin classifier for few-shot learning (FS-TMMC). Our method utilizes the unlabeled query examples for transductive inference. Different from the classical maximum margin classifier, FS-TMMC also assigns labels to the query examples when constructing the classifier with the largest geometric margin. By increasing the weight of the loss function evaluated at the examples in the query set step by step, we gradually adjust the separating hyperplane of the maximum margin classifier. We conducte extensive experiments on the few-shot learning benchmarks including miniImageNet, tieredImageNet and CUB, where we achieve state-of-the-art results. 

\bibliographystyle{unsrt}
\bibliography{references}
\end{document}